\def\year{2019}\relax
\documentclass[letterpaper]{article} 
\usepackage{aaai19}  
\usepackage{times}  
\usepackage{helvet}  
\usepackage{courier}  
\usepackage{url}  
\usepackage{graphicx}  

\usepackage{mathptmx}
\usepackage{amssymb}
\usepackage{amsmath}
\usepackage{amsfonts}
\usepackage{hyphenat}
\usepackage{color}
\usepackage{textcomp}
\usepackage{enumitem}
\usepackage{xspace}
\usepackage{subfigure}
\usepackage{tikz-cd}
\usepackage{numprint}

\DeclareSymbolFont{greek}{OML}{cmm}{m}{n}
\DeclareMathSymbol{\alpha}{\mathalpha}{greek}{"0B}
\DeclareMathSymbol{\beta}{\mathalpha}{greek}{"0C}
\DeclareMathSymbol{\gamma}{\mathalpha}{greek}{"0D}
\DeclareMathSymbol{\delta}{\mathalpha}{greek}{"0E}
\DeclareMathSymbol{\epsilon}{\mathalpha}{greek}{"0F}
\DeclareMathSymbol{\zeta}{\mathalpha}{greek}{"10}
\DeclareMathSymbol{\eta}{\mathalpha}{greek}{"11}
\DeclareMathSymbol{\theta}{\mathalpha}{greek}{"12}
\DeclareMathSymbol{\iota}{\mathalpha}{greek}{"13}
\DeclareMathSymbol{\kappa}{\mathalpha}{greek}{"14}
\DeclareMathSymbol{\lambda}{\mathalpha}{greek}{"15}
\DeclareMathSymbol{\mu}{\mathalpha}{greek}{"16}
\DeclareMathSymbol{\nu}{\mathalpha}{greek}{"17}
\DeclareMathSymbol{\xi}{\mathalpha}{greek}{"18}
\DeclareMathSymbol{\pi}{\mathalpha}{greek}{"19}
\DeclareMathSymbol{\rho}{\mathalpha}{greek}{"1A}
\DeclareMathSymbol{\sigma}{\mathalpha}{greek}{"1B}
\DeclareMathSymbol{\tau}{\mathalpha}{greek}{"1C}
\DeclareMathSymbol{\upsilon}{\mathalpha}{greek}{"1D}
\DeclareMathSymbol{\phi}{\mathalpha}{greek}{"1E}
\DeclareMathSymbol{\chi}{\mathalpha}{greek}{"1F}
\DeclareMathSymbol{\psi}{\mathalpha}{greek}{"20}
\DeclareMathSymbol{\omega}{\mathalpha}{greek}{"21}
\DeclareMathSymbol{\varepsilon}{\mathalpha}{greek}{"22}
\DeclareMathSymbol{\vartheta}{\mathalpha}{greek}{"23}
\DeclareMathSymbol{\varpi}{\mathalpha}{greek}{"24}
\DeclareMathSymbol{\varrho}{\mathalpha}{greek}{"25}
\DeclareMathSymbol{\varsigma}{\mathalpha}{greek}{"26}
\DeclareMathSymbol{\varphi}{\mathalpha}{greek}{"27}
\DeclareSymbolFont{otone}{OT1}{cmr}{m}{n}
\DeclareMathSymbol{\Gamma}{\mathalpha}{otone}{0}
\DeclareMathSymbol{\Delta}{\mathalpha}{otone}{1}
\DeclareMathSymbol{\Theta}{\mathalpha}{otone}{2}
\DeclareMathSymbol{\Lambda}{\mathalpha}{otone}{3}
\DeclareMathSymbol{\Xi}{\mathalpha}{otone}{4}
\DeclareMathSymbol{\Pi}{\mathalpha}{otone}{5}
\DeclareMathSymbol{\Sigma}{\mathalpha}{otone}{6}
\DeclareMathSymbol{\Upsilon}{\mathalpha}{otone}{7}
\DeclareMathSymbol{\Phi}{\mathalpha}{otone}{8}
\DeclareMathSymbol{\Psi}{\mathalpha}{otone}{9}
\DeclareMathSymbol{\Omega}{\mathalpha}{otone}{10}
\DeclareSymbolFont{syms}{OML}{cmm}{m}{it}
\DeclareMathSymbol{\partial}{\mathord}{syms}{"40}
\DeclareMathAlphabet{\mathbold}{OML}{cmm}{b}{it}
\DeclareSymbolFont{largesymbols}{OMX}{cmex}{m}{n}

\newcommand{\hide}[1]{}

\newcommand{\xhdr}[1]{\vspace{1.7mm}\noindent{{\bf #1.}}}
\newcommand{\xhdrNoPeriod}[1]{\vspace{1.7mm}\noindent{{\bf #1}}}

\newcommand{\ie}{{i.e.}\xspace}
\newcommand{\eg}{{e.g.}\xspace}
\newcommand{\cf}{{cf.}\xspace}

\newcommand{\vs}{{vs.}\xspace}
\newcommand{\etc}{{etc.}\xspace}

\newcommand{\Secref}[1]{Sec.~\ref{#1}}
\newcommand{\Eqnref}[1]{Eq.~\ref{#1}}

\newcommand{\Tabref}[1]{Table~\ref{#1}}
\newcommand{\Figref}[1]{Fig.~\ref{#1}}

\newcommand{\hdln}[1]{\textit{#1}}

\newcommand{\exampleNum}[1]{H#1}

\newcommand{\Unfun}{\textit{Unfun.me}}
\newcommand{\UnfunTheHeadline}{\textit{Unfun the headline!\xspace}}
\newcommand{\RealOrNot}{\textit{Real or not?\xspace}}
\newcommand{\sbsl}{similar\hyp but\hyp serious\hyp looking\xspace}

\title{
Reverse-Engineering Satire, or\\
``Paper on Computational Humor Accepted despite Making Serious Advances''
}

\author{
Robert West%
\thanks{Research done partly at Microsoft Research.}\\
EPFL\\
robert.west@epf\/l.ch
\And
Eric Horvitz\\
Microsoft Research\\
horvitz@microsoft.com
}

\begin{document}

\maketitle

\begin{abstract}
Humor is an essential human trait. Efforts to understand humor have called out links between humor and the foundations of cognition, as well as the importance of humor in social engagement. As such, it is a promising and important subject of study, with relevance for artificial intelligence and human--computer interaction.
Previous computational work on humor has mostly operated at a coarse level of granularity, \eg, predicting whether an entire sentence, paragraph, document, \etc, is humorous.
As a step toward deep understanding of humor, we seek fine\hyp grained models of attributes that make a given text humorous.
Starting from the observation that satirical news headlines tend to resemble serious news headlines, we build and analyze a corpus of satirical headlines paired with nearly identical but serious headlines.
The corpus is constructed via \Unfun, an online game that incentivizes players to make minimal edits to satirical headlines with the goal of making other players believe the results are serious headlines.
The edit operations used to successfully remove humor pinpoint the words and concepts that play a key role in making the original, satirical headline funny.
Our analysis reveals that the humor tends to reside toward the end of headlines, and primarily in noun phrases, and that most satirical headlines follow a certain logical pattern, which we term {\em false analogy}.
Overall, this paper deepens our understanding of the syntactic and semantic structure of satirical news headlines and provides insights for building humor\hyp producing systems.
\end{abstract}

\section{Introduction}
\label{sec:intro}

Humor is a uniquely human trait that plays an essential role in our everyday lives and interactions.
Psychologists have pointed out the role of humor in human cognition, including its link to the identification of surprising connections in learning and problem solving, as well as the importance of humor in social engagement \cite{martin2010psychology}.
Humor is a promising area for studies of intelligence and its automation:
it is hard to imagine a computer passing a rich Turing test without being able to understand and produce humor.
As computers increasingly take on conversational tasks (\eg, in chat bots and personal assistants), the ability to interact with users naturally is gaining importance, but human--computer interactions will never be truly natural without giving users the option to say something funny and have it understood that way; \eg, recent work has shown that misunderstanding of playful quips can be the source of failures in conversational dialog in open-world interaction \cite{andrist2016you}.

Given how tied humor is to the human condition, the phenomenon has challenged some of the greatest thinkers throughout history and has been the subject of much academic research across over 20 disciplines \cite{raskin2008primer}, including computer science \cite{binsted2006computational}, where researchers have developed algorithms for detecting, analyzing, and generating humorous utterances (\cf\ \Secref{sec:relwork}).

\begin{figure}[t]
  \includegraphics[width=\columnwidth]{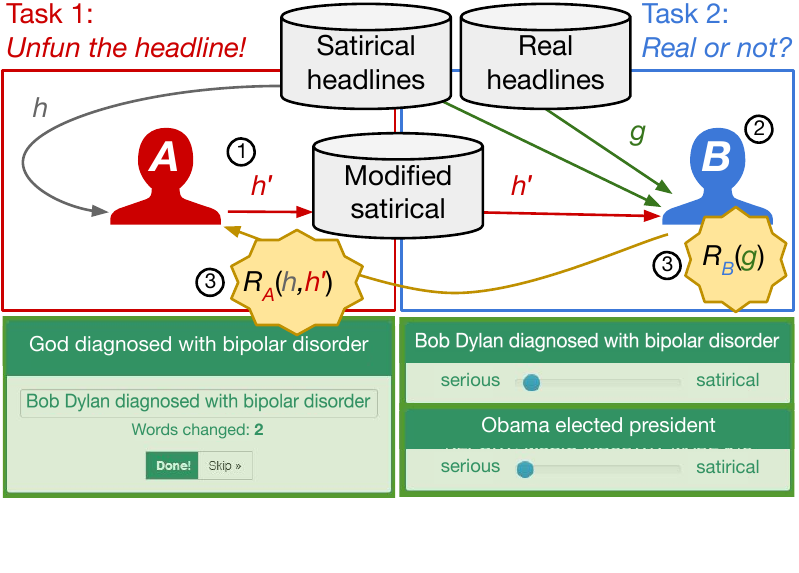}
  \vspace{-14mm}
  \caption{
  \Unfun, a game for building a
  corpus of pairs $(h,h')$ of satirical and \sbsl headlines.
  Numbers: order of steps.
  Screenshots: running example
  ($h=$ \hdln{God diagnosed with bipolar disorder;}
  $h'=$ \hdln{Bob Dylan diagnosed with bipolar disorder;}
  $g=$ \hdln{Obama elected president}).
  }
  \label{fig:game_diagram}
  \vspace{-4mm}
\end{figure}

The automated analysis of humor is complicated by the fact that most humorous texts have a complex narrative structure that is difficult to disentangle;
\eg, typical jokes---the type of humorous text studied most in the literature---carefully set the stage to build certain expectations in the audience, which are then turned upside down in the punchline.
To circumvent the difficulties imposed by narrative structure, we focus on a specific humorous genre: satirical news.
Satirical news articles, on the surface, mimic the format typical of mainstream journalism, but unlike serious news articles, they do not aim to relate facts, but rather to ridicule individuals, groups, or society.
Crucially, though, satirical news stories are typically written headline\hyp first:
only if the headline is funny in and of itself is the rest of the story written \cite{thisamericanlife}.
This is markedly different from real news stories and means that \textbf{satirical news headlines} can be studied in isolation from the full stories, whose essence they convey in a concise form
with minimal narrative structure.

An additional advantage of satirical headlines is that they mimic the formulaic style of serious news headlines, which limits their syntactic variability and allows us to better control for syntax and focus on semantics.
Moreover, satirical headlines are similar to serious news headlines not only in style but also in content:
changing a single word often suffices to make a satirical headline sound like serious news.

\xhdr{Running example}
For instance, changing \textit{God} to \textit{Bob Dylan} turns the satirical headline
\textbf{\hdln{God diagnosed with bipolar disorder,}}
which was published in the satirical newspaper \textit{The Onion}, into
\textbf{\hdln{Bob Dylan diagnosed with bipolar disorder,}}
which could appear \textit{verbatim} in a serious newspaper.

A large corpus of such pairs of satirical and \sbsl headlines would open up exciting opportunities for humor research.
For instance, it would allow us to understand why a satirical text is funny at a finer granularity than previously possible, by identifying the exact words that make the difference between serious and funny.
This is a striking difference from most previous research, where usually the \textit{average} satirical headline is compared to the \textit{average} serious one \cite{mihalcea2007characterizing}.
Moreover, while the principal goal of this research has been to achieve new insights about humor, we also imagine new applications.
For example, if we attained a grasp on the precise differences between satirical and serious headlines, we might be able to create procedures for transforming real news headlines into satirical headlines with minimal changes.

To create an aligned corpus, a first idea would be to automatically pair satirical with serious news headlines: start with a satirical headline and find the most similar serious headline written around the same time.
It is hard to imagine, though, that this process would yield many pairs of high lexical and syntactic similarity.
An alternative idea would be to use crowdsourcing: show serious headlines to humans and ask them to turn them into satirical headlines via minimal edits.
Unfortunately, this task requires a level of creative talent that few people have. Even at \textit{The Onion,} America's most prominent satirical newspaper, only 16 of 600 headlines generated each week (less than 3\%) are accepted \cite{thisamericanlife}.

The crucial observation is that the task is much easier in the reverse direction:
it is typically straightforward to \textbf{remove the humor} from a satirical headline by applying small edits that turn the headline into one that looks serious and could conceivably be published in a real news outlet.
In other words, reversing the creative effort that others have already invested in crafting a humorous headline requires much less creativity than crafting the headline in the first place.
We thus adopt this reverse\hyp crowdsourcing approach, by designing a \textbf{game with a purpose} \cite{vonahn+dabbish2008}.

The game is called \textbf{\Unfun{}} and is described graphically in \Figref{fig:game_diagram}.
A player $A$ of the game is given a satirical news headline $h$ and asked to modify it in order to fool other players into believing that the result $h'$ is a real headline from a serious news outlet.
The reward $R_A(h,h')$ received by the player $A$ who modified the satirical headline
increases with the fraction of other players rating the modified headline $h'$ as serious and
decreases with the number of words changed in the original headline $h$.

\xhdr{Contributions}
Our main contributions are twofold.
First, we present \Unfun, an online game for collecting a corpus of pairs of satirical news headlines aligned to \sbsl headlines (\Secref{sec:game}).
Second, our analysis of these pairs (\Secref{sec:game dynamics}--\ref{sec:semantic analysis}) reveals key properties of satirical headlines at a much finer level of granularity than prior work (\Secref{sec:relwork}).
Syntactically (\Secref{sec:syntactic analysis}), we conclude that the humor tends to reside in noun phrases, and with increased likelihood toward the end of headlines, giving rise to what we term ``micro\hyp punchlines''.
Semantically (\Secref{sec:semantic analysis}), we observe that original and modified headlines are usually opposed to each other along certain dimensions crucial to the human condition (\eg, \textit{high \vs low stature,} \textit{life \vs death}), and that satirical headlines are overwhelmingly constructed according to a \textit{false\hyp analogy} pattern.
We conclude the paper by discussing our findings in the context of established theories of humor (\Secref{sec:discussion}).

\section{Game description: \Unfun}
\label{sec:game}

Here we introduce \Unfun, our game for collecting pairs of satirical and similar\hyp but\hyp serious\hyp looking headlines.
The game, available online at \url{http://unfun.me} and visually depicted in \Figref{fig:game_diagram}, challenges players in two tasks.


\xhdrNoPeriod{Task 1: \UnfunTheHeadline{}}
This is the core task where the reverse\hyp engineering of satire happens (left panel in \Figref{fig:game_diagram}).
A player, $A$, is given a satirical headline $h$ and is asked to turn it into a headline $h'$ that could conceivably have been published by a serious news outlet, by changing as few words as possible.

\xhdrNoPeriod{Task 2: \RealOrNot{}}
Whether on purpose or not, player $A$ may have done a bad job in task~1, and $h'$ may still be humorous.
Detecting and filtering such cases is the purpose of task~2 (right panel in \Figref{fig:game_diagram}), where $h'$ is shown to another player, $B$, who is asked to indicate her belief $p_B(h')$ that $h'$ comes from a serious news outlet using a slider bar ranging from 0\% to 100\%.
We shall refer to $p_B(h')$ as $B$'s \textbf{seriousness rating} of $h'$.
For reasons that will become clear below, player $B$ also indicates her belief $p_B(g)$ for a second, unmodified headline $g$ (unrelated to $h$) that originates from either a serious or a satirical news outlet.
The two headlines $h'$ and $g$ are presented in random order, in order to avoid biases.

For the purpose of incentivizing players to make high\hyp quality contributions, we reward them as follows.

\xhdr{Reward for task 1}
As player $A$ is supposed to \textit{remove the humor} from $h$ via a \textit{minimal modification,} his reward $R_A(h,h')$ increases
(1)~with the average rating $r(h')$ that the modified headline $h'$ receives from all $n$ players $B_1, \dots, B_n$ who rate it and
(2)~with the similarity $s(h,h')$ of $h$ and $h'$:
\begin{equation}
R_A(h,h') = \sqrt{r(h') \; s(h,h')} \;,
\label{eqn:R_A}
\end{equation}
\begin{equation*}
\text{where} 
\;\;\;\;
r(h') = \frac{1}{n} \sum_{i=1}^n p_{B_i}(h'),
\;\;\;\;
s(h,h') = 1-\frac{d(h,h')}{\max\{|h|,|h'|\}},
\end{equation*}
where, in turn, $|x|$ is the number of tokens (\ie, words) in a string $x$,
and $d(h,h')$, the \textbf{token\hyp based edit distance} \cite{navarro2001guided} between $h$ and $h'$, \ie, the minimum number of insertions, deletions, and substitutions by which $h$ can be transformed into $h'$, considering as the basic units of a string its tokens, rather than its characters.
The geometric mean was chosen in \Eqnref{eqn:R_A} because it is zero whenever one of the two factors is zero (which is not true for the more standard arithmetic mean):
a modified headline that seems very serious, but has nothing to do with the original, should not receive any points, nor should a headline that is nearly identical to the original, but retains all its humor.

\xhdr{Reward for task 2}
Since player $B$'s very purpose is to determine whether $h'$ is without humor, we do not have a ground\hyp truth rating for $h'$.
In order to still be able to reward player $B$ for participating in task~2, and to incentivize her to indicate her true opinion about $h'$,
we also ask her for her belief $p_B(g)$ regarding a headline $g$ for which we do have the ground truth of ``serious'' \vs ``satirical''.
The reward $R_B(g)$ that player $B$ receives for rating headline $g$ is then
\begin{equation}
R_B(g) =        \begin{cases}
                \log(p_B(g)) & \text{if $g$ is serious,}\\
                \log(1-p_B(g)) & \text{if $g$ is satirical.}\\
                \end{cases}
\label{eqn:R_B}
\end{equation}
Note that this is a \textit{proper scoring rule} \cite{gneiting2007strictly}, \ie, player $B$ maximizes her expected reward by indicating her true belief.
This would not be true for the more straightforward scoring formula without logarithms,
which would drive players to report beliefs of 0 or 1 instead of their true beliefs.
Also, as $h'$ and $g$ are shown in random order, $B$ does not know which is which, and her optimal strategy is to indicate her true belief on both.

\xhdr{Overall game flow}
Whenever a user wants to play, we generate a type-1 task with probability $\alpha=1/3$ and a type-2 task with probability $1-\alpha=2/3$, such that we can collect two ratings per modified headline.
As mentioned, ratings from task~2 can serve as a filter, and we can increase its precision at will by decreasing $\alpha$.
To make rewards more intuitive and give more weight to the core task~1, we translate and scale rewards such that $R_A(\cdot,\cdot) \in [0, 1000]$ and $R_B(\cdot) \in [0, 200]$.
We also implemented additional incentive mechanisms such as badges, high-score tables, and immediate rewards for participating, but we omit the details for space reasons.

\xhdr{Satirical and serious headlines}
The game requires corpora of satirical as well as serious news headlines as input.
Our satirical corpus consists of 9,159 headlines published by the well\hyp known satirical newspaper \textit{The Onion;}
our serious corpus, of 9,000 headlines drawn from 9 major news websites.

\xhdr{Data and code}
We make the data collected via \Unfun, as well as our code for analyzing it, publicly available online \cite{github}.

\section{Analysis of game dynamics}
\label{sec:game dynamics}

\begin{figure*}[t]
  \centering
  \subfigure[\hspace*{8mm}]{
    \hspace{-4mm}
    \includegraphics[scale=1]{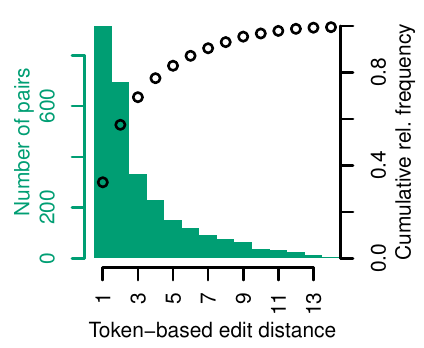}
    \label{fig:edit_dist_hist_TOKENS_ALL}
  }
  \subfigure[\hspace*{-7mm}]{
    \hspace{-2mm}
    \includegraphics[scale=1]{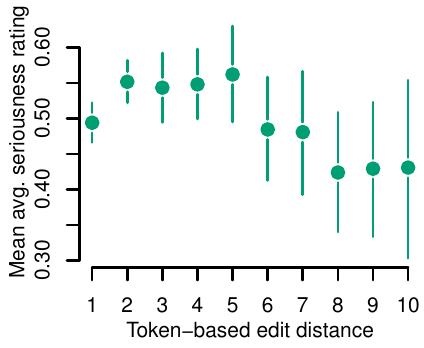}
    \label{fig:edit_dist_vs_rating}
  }
  \subfigure[\hspace*{-9mm}]{
    \hspace{-2mm}
    \includegraphics[scale=1]{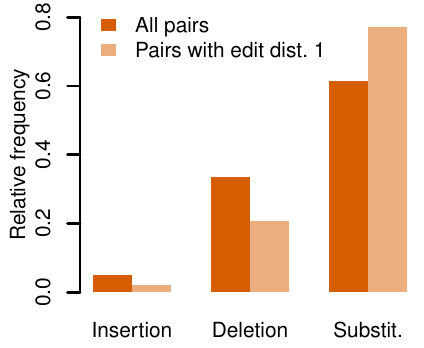}
    \label{fig:edit_type_hist_SUCCESSFUL}
  }
  \subfigure[\hspace*{-1mm}]{
    \hspace{-2mm}
    \includegraphics[scale=1]{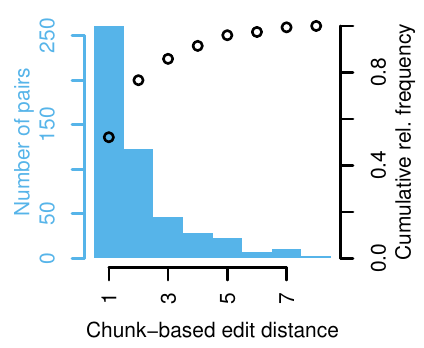}
    \label{fig:edit_dist_hist_CHUNKS}
  }
  \vspace*{-4mm}
  \caption{
  \textbf{(a)} Distribution of token\hyp based edit distance in headline pairs collected via \Unfun.
  \textbf{(b)} Tradeoff of edit distance \vs seriousness rating (only pairs with at least 2 ratings; with bootstrapped 95\% confidence intervals).
  \textbf{(c)} Distribution of token\hyp based edit operations (successful pairs only, \cf \Secref{sec:game dynamics}).
  \textbf{(d)} Distribution of chunk\hyp based edit distance (successful pairs only).
  }
  \label{fig:edit_properties_TOKENS}
  \vspace{-4mm}
\end{figure*}

Via \Unfun,
we have collected 2,801 modified versions $h'$ for 1,191 distinct satirical headlines $h$
(2.4 pairs per satirical headline).
All but 7 modified headlines have received at least one rating, and 1,806 (64\%), at least two (mean\slash median: 2 ratings per modified headline).
The modified headlines (ratings) came from 582 (546) unique user ids (mean\slash median: 4.8\slash 2 modified headlines per user; 10\slash 4 ratings per user).

We start by analyzing the edit operations players perform in task~1 and the seriousness ratings they provide in task~2.
The main objects of study are pairs $(h,h')$ consisting of an original satirical headline $h$ and a modified version $h'$, which we shall simply call \textbf{pairs} in what follows.

\xhdr{Edit distance}
The first interesting question is how much players tend to modify original satirical headlines $h$ in order to expunge the humor from them.
We quantify this notion via the token\hyp based edit distance $d(h,h')$ between the satirical headline $h$ and the modified version $h'$ (\cf \Secref{sec:game}).
\Figref{fig:edit_dist_hist_TOKENS_ALL}, which plots the distribution of edit distance, shows that very small edits are most common, as incentivized by the reward structure of the game (\Eqnref{eqn:R_A}).
In particular, 33\% of all pairs have the smallest possible edit distance of 1, and 57\% (69\%) have a distance up to 2 (3).

\xhdr{Tradeoff of edit distance \vs seriousness rating}
The reward structure of the game (\Eqnref{eqn:R_A}) does not, however, exclusively encourage small edits.
Rather, there is a tradeoff:
larger edits (bad) make it easier to remove the humor (good),
while smaller edits (good) run the risk of not fully removing the humor (bad).
\Figref{fig:edit_dist_vs_rating}, which plots the mean average seriousness rating $r(h')$ of modified headlines $h'$ as a function of the edit distance $d(h,h')$,
shows how this tradeoff plays out in practice.
For edit distances between 1 and 5 (83\% of all pairs, \cf \Figref{fig:edit_dist_hist_TOKENS_ALL}), seriousness ratings correlate positively with edit distance.
In particular, it seems harder to remove the humor by changing one word than by changing two words, whereas the marginal effect is negligible when allowing for even larger edits.
The positive correlation does not hold for the much smaller number (17\%) of pairs with an edit distance above 5.
Inspecting the data, we find that this is caused by headlines so inherently absurd that even large edits cannot manage to remove the humor from them.

\xhdr{Seriousness ratings}
Recall that, in task~2, players attribute seriousness ratings to modified headlines $h'$, as well as to unmodified serious or satirical headlines $g$.
We find that, in all three cases, the distribution of seriousness ratings is bimodal, with extreme values close to 0 or 1 being most common.
Hence, we binarize ratings into two levels, ``satirical'' (rating below 0.5) and ``serious'' (rating above 0.5).

In order to see how people rate serious, satirical, and modified headlines, respectively,
\Tabref{tbl:confusion_matrix} aggregates ratings by headline (considering only the 1,806 headlines with at least two ratings) and splits the headlines into three groups:
``consensus serious'' (over 50\% ``serious'' ratings),
``no consensus'' (exactly 50\%), and
``consensus satirical'' (under 50\%).

We make two observations. First, modified headlines $h'$ (column~3 of \Tabref{tbl:confusion_matrix}) are distributed roughly evenly over the three groups; \ie, there are about as many headlines from which the humor has been successfully removed (``consensus serious'') as not (``consensus satirical'').
The most useful modified headlines for our purposes are those from the ``consensus serious'' group, as they likely do not carry the humor of the original $h$ anymore.
Hence, we shall restrict our subsequent analyses to the corresponding 654 \textbf{successful pairs.}%
\footnote{
\label{footnote:edit_dist_hist_TOKENS_ALL}
As a sanity check, we computed the edit\hyp distance distribution for successful pairs only, finding no big differences from \Figref{fig:edit_dist_hist_TOKENS_ALL}.
}
Second, the ratings are heavily skewed toward the ground truth for unmodified serious (column~1) and satirical (column~2) headlines; \ie, players can typically well distinguish serious from satirical headlines (but \cf discussion in \Secref{sec:discussion}).

\begin{table}[t]
\centering
\vspace{-2.5mm}
\caption{
Rating distributions for pairs with at least 2 ratings.
``No consensus'' large as most pairs have exactly 2 ratings.
}
\label{tbl:confusion_matrix}
{\small
\begin{tabular}{llll}
  \hline
\textbf{Aggregate rating} & \textbf{Serious} & \textbf{Satirical} & \textbf{Modified} \\
  \hline
Consensus serious    & 777 (57\%) & 105  (8\%) & 654 (36\%) \\
No consensus         & 447 (33\%) & 368 (27\%) & 570 (32\%) \\
Consensus satirical  & 133 (10\%) & 871 (65\%) & 582 (32\%) \\
  \hline
\end{tabular}
}
\vspace{-4mm}
\end{table}

\xhdr{Insertions, deletions, substitutions}
When computing the edit distance $d(h,h')$ using dynamic programming, we can also keep track of an optimal sequence of edit operations (insertions, deletions, substitutions) for transforming $h$ into $h'$ \cite{navarro2001guided}.
In \Figref{fig:edit_type_hist_SUCCESSFUL}, we plot the distribution of edit operations, macro\hyp averaged over all pairs.
We see that substitutions clearly dominate (61\%), followed by deletions (34\%), with insertions being very rare (5\%).

Pairs with edit distance 1 are particularly interesting, as they are the most similar, as well as the most frequent
(\Figref{fig:edit_dist_hist_TOKENS_ALL}, footnote \ref{footnote:edit_dist_hist_TOKENS_ALL}).
Also, the optimal edit sequence may not be unique in general, but for edit distance 1 it is.
Hence, \Figref{fig:edit_type_hist_SUCCESSFUL} also displays the distribution over edit operations for pairs with edit distance 1 only.
Here, substitutions dominate even more (77\%), and insertions are even rarer (2\%).

Reversing the direction of the editing process, we hence conclude that writers of satirical headlines tend to work overwhelmingly by substituting words in (hypothetical) similar\hyp but\hyp serious headlines, and to a certain degree by adding words, but very rarely by deleting words.

\section{Syntactic analysis of aligned corpus}
\label{sec:syntactic analysis}

Next, we go one level deeper and ask: what parts of a satirical headline should be modified in order to remove the humor from it, or conversely,
what parts of a serious headline should be modified in order to add humor?
We first tackle this question from a syntactic perspective,
before moving to a deeper, semantic perspective in \Secref{sec:semantic analysis}.

\xhdr{From tokens to chunks}
We analyze syntax at an intermediate level of abstraction between simple sequences of part-of-speech (POS) tags and complex parse trees, by relying on a \textit{chunker} (also called \textit{shallow parser}).
We use OpenNLP's maximum entropy chunker \cite{berger1996maximum}, after retraining it to better handle pithy, headline\hyp style text.
The chunker takes POS\hyp tagged text as input and groups subsequent tokens into meaningful phrases \textbf{(chunks)} without inferring the recursive structure of parse trees;
\eg,
our running example (\Secref{sec:intro}) is chunked as
[NP \hdln{Bob Dylan}] [VP \hdln{diagnosed}] [PP \hdln{with}] [NP \hdln{bipolar disorder}]
(chunk labels expanded in \Tabref{tbl:histogram_of_modified_phrases}).
Chunks are handy because they abstract away low\hyp level details; \eg, changing \hdln{God} to \hdln{Bob Dylan} requires a token\hyp based edit distance of 2, but a chunk\hyp based distance of only 1, where the latter is more desirable because it more closely captures the conceptual modification of one entity being replaced by another entity.

Chunking all 9,159 original headlines from our \textit{The Onion} corpus,
we find the most frequent chunk pattern to be
NP VP NP PP NP (4.8\%; \eg, \exampleNum{2} in \Figref{fig:examples_and_diagrams}(a)),
followed by NP VP NP (4.3\%; \eg, \exampleNum{4}) and
NP VP PP NP (3.3\%; \eg, \exampleNum{9}).

To control for syntactic effects, it is useful to study a large number of pairs $(h,h')$ where all original headlines $h$ follow a fixed syntactic pattern.
We therefore gave priority to headlines of the most frequent pattern (NP VP NP PP NP) for a certain time period when sampling satirical headlines as input to task~1,
such that, out of all 2,801 $(h,h')$ pairs collected in task~1,
$h$ follows that pattern in 21\% of all cases.

\xhdr{Chunk-based edit distance}
Recomputing edit distances at the chunk level, rather than the token level, we obtain the chunk\hyp based edit distance distribution of \Figref{fig:edit_dist_hist_CHUNKS}.
It resembles the token\hyp based edit distance distribution of \Figref{fig:edit_dist_hist_TOKENS_ALL},
with the difference that the smallest possible distance of 1 is even more prevalent (52\% \vs 33\% of pairs), due to the fact that modifying a single chunk frequently corresponds to modifying multiple tokens.
Since, moreover, the vast majority (97\%) of all single\hyp chunk edits are substitutions, we now focus on 254 $(h,h')$ pairs where exactly one chunk of $h$ has been modified (henceforth \textbf{single\hyp substitution pairs}).
This accounts for about half of all successful pairs (after discarding pairs that were problematic for the chunker).

\xhdr{Dominance of noun phrases}
We now ask which syntactic chunk types (noun phrases, verb phrases, \etc) are modified to remove humor.
In doing so, we need to be careful, as some chunk types are more common \textit{a priori} than others;
\eg, 59\% of all chunks in original satirical headlines are noun phrases, 20\%, verb phrases, \etc
We therefore compare the empirical distribution of modified chunks with this prior distribution, via the ratio of the two (termed \textit{lift}).
\Tabref{tbl:histogram_of_modified_phrases} shows that noun phrases constitute 89\% of the modified chunks (lift 1.52), whereas all other chunk types are less frequent than under the prior.
We conclude that the humor of satirical news headlines tends to reside in noun phrases.

\begin{table}[t]
\centering
\vspace{-2.5mm}
\caption{
Distribution of syntactic chunk types in single\hyp substitution pairs (only showing types modified at least once).
}
\label{tbl:histogram_of_modified_phrases}
\npdecimalsign{.}
\nprounddigits{2}
{\small
\begin{tabular}{lln{2}{2}n{2}{2}n{1}{2}}
\hline
\textbf{Label} & \textbf{Chunk type} & \textbf{Modified} & \textbf{Prior} & \textbf{Lift}\\
\hline
NP   & Noun phrase & 89.3700787\% & 58.6279840\% & 1.52435872\\
VP   & Verb phrase & 9.4488189\% & 20.1520122\% & 0.46887719\\
ADJP & Adjective phrase & 0.7874016\% & 1.4899700\% & 0.52846807\\
PP   & Preposition & 0.3937008\% & 17.3957943\% & 0.02263195\\
\hline
\end{tabular}
}
\npnoround
\end{table}

\begin{figure}[t]
  \centering
  \hspace*{-6mm}
  \includegraphics[scale=0.5]{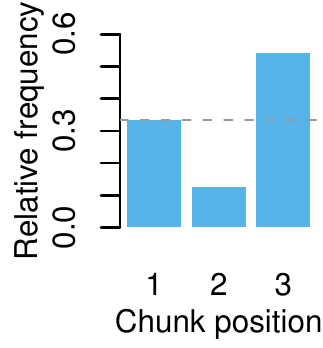}
  \hspace*{-3mm}
  \includegraphics[scale=0.5]{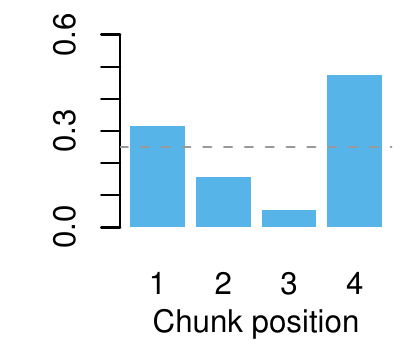}
  \hspace*{-3mm}
  \includegraphics[scale=0.5]{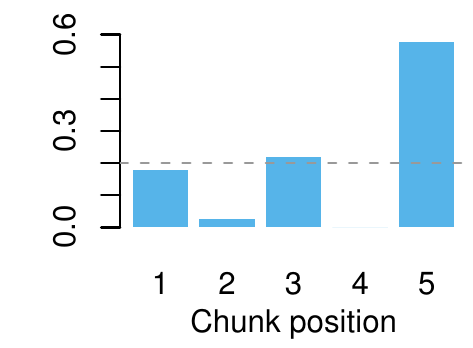}
  \hspace*{-3mm}
  \includegraphics[scale=0.5]{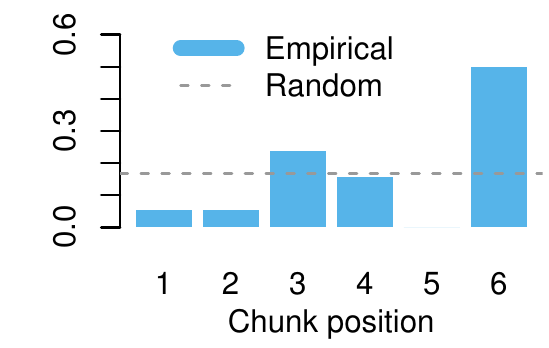}
  \hspace*{-5mm}
  \vspace{-4mm}
  \caption{
  Distributions of modified chunk positions in single\hyp substitution pairs, for original headlines containing 3 to 6 chunks (number of pairs for each length: 24, 38, 123, 38).
  }
  \label{fig:edit_pos_distrib}
  \vspace{-3mm}
\end{figure}

\xhdr{Micro-punchlines}
We now ask where in terms of location within a headline the humor tends to reside.
To answer this question, we compute the position of the modified chunk in each headline's chunk sequence and plot the distribution of modified positions in \Figref{fig:edit_pos_distrib}.
We see that, regardless of headline length, modifications to the last chunk are particularly overrepresented.%
\footnote{
We ascertained that the effect is not due to trailing chunks potentially being
(1)~longer and
(2)~more likely to be noun phrases.
}
This is an important finding:
we have previously (\Secref{sec:intro}) argued that satirical headlines consist of a punchline only, with minimal narrative structure, and indeed it was this very intuition that led us to investigate headlines in isolation.
Given \Figref{fig:edit_pos_distrib}, we need to revise this statement slightly:
although satirical headlines consist of a single sentence, they are often structured---at a micro\hyp level---akin to more narrative jokes, where the humorous effect also comes with the very last words.
Put differently, the final words of satirical headlines often serve as a ``micro\hyp punchline''.%
\footnote{
Strictly speaking, the findings of \Secref{sec:syntactic analysis} only pertain to satirical headlines that are already similar to hypothetical serious headlines, due to
selection bias (we only study headlines that players of \Unfun{} chose to modify)
and due to our focus on single\hyp substitution pairs (about 50\% of successful pairs).
}

\section{Semantic analysis of aligned corpus}
\label{sec:semantic analysis}

After characterizing aligned pairs syntactically, we now move to the semantic level.
We first analyze the aligned pairs obtained from \Unfun{} and later discuss our findings in the broader context of established theories of humor (\Secref{sec:discussion}).

\xhdr{Example}
Before a more general analysis, let us first consider again our running example (\Secref{sec:intro}), \hdln{God diagnosed with bipolar disorder}.
This satirical headline works by blending two realms that are fundamentally opposed---the human and the divine---by talking about God as a human.
Although the literally described situation is impossible (God is perfect and cannot possibly have a disease), the line still makes sense by expressing a crucial commonality between bipolar humans and God, namely that both may act unpredictably.
But for humans, being unpredictable (due to bipolarity) is a sign of imperfection, whereas for God it is a sign of perfection (``The Lord moves in mysterious ways''), and it is this opposition
that makes the line humorous.

The main advantage of our aligned corpus is that it lets us generalize this \textit{ad-hoc} analysis of a particular example to a large and representative set of satirical headlines by pinpointing the essential, humor\hyp carrying words in every headline:
if the humor has been successfully removed from a headline $h$ by altering certain words, then we know that these very words are key to making $h$ funny.

This is especially true for single\hyp substitution pairs;
\eg, in the running example, \hdln{God} was replaced by \hdln{Bob Dylan} (a particular human), giving rise to the serious\hyp sounding
\hdln{Bob Dylan diagnosed with bipolar disorder}.
The automatically extracted chunk pair \{\hdln{God}, \hdln{Bob Dylan}\} surfaces both the crucial commonality in the context of the headline (unpredictability) and the crucial opposition (God \vs human; unpredictability as a good \vs bad trait).

While the semantic analysis of original \vs substituted chunks may be difficult to automate, having access to explicit chunk pairs tremendously facilitates a large\hyp scale human analysis.
Conducting such an analysis revealed that the above pattern of a crucial commonality combined with a crucial opposition occurs in a large fraction of satirical headlines, and particularly in nearly all single\hyp substitution pairs.

\begin{figure*}
\includegraphics{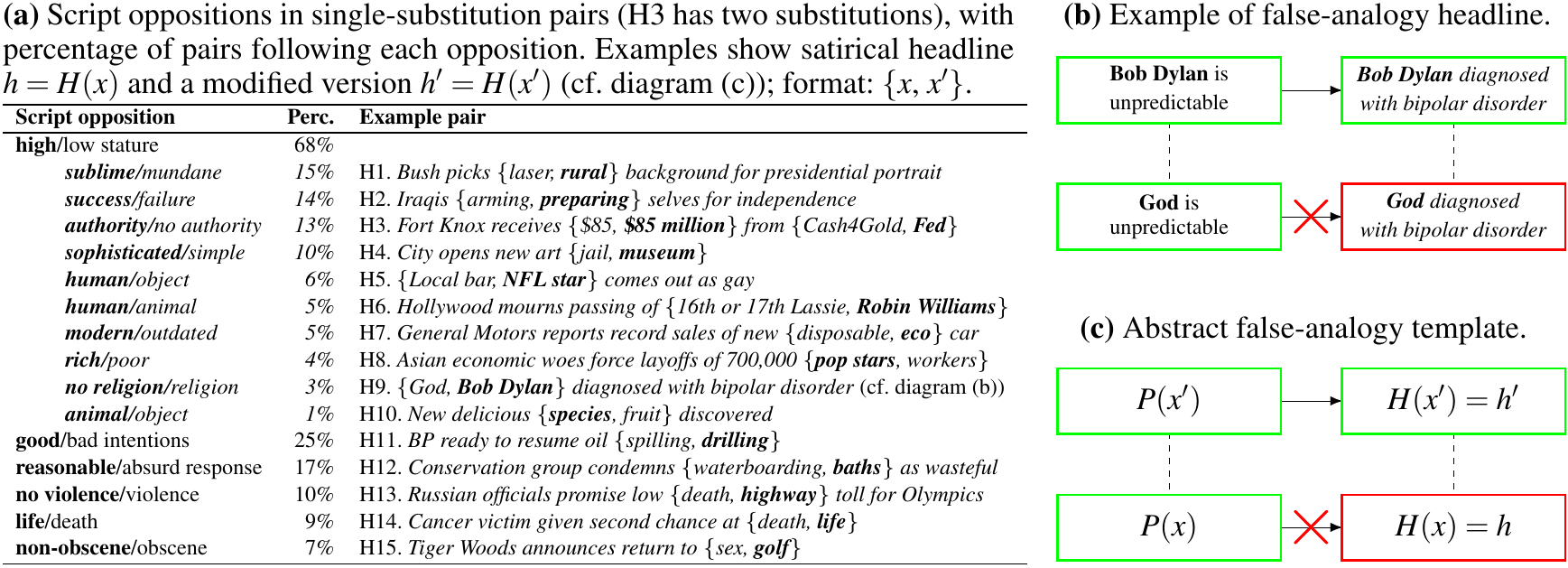}
\vspace{-3mm}
\caption{
\textbf{(a)}~Script oppositions and examples.
\textbf{(b)}~Example of false-analogy headline.
\textbf{(c)}~Abstract false-analogy template.
}
\label{fig:examples_and_diagrams}
\vspace{-1mm}
\end{figure*}

\xhdr{Script opposition}
The crucial opposition has been called \textit{script opposition} by humor theorists (\cf \Secref{sec:discussion}), and we henceforth adopt the same term.
Inspecting all 254 single\hyp substitution pairs, we found each pair to be in at least one of 6 oppositions, all representing ``good''-\vs-``bad'' dichotomies that are essential to the human condition, such as
\textit{high\slash low stature},
\textit{life\slash death,} or
\textit{non\hyp obscene\slash obscene.}
All 6 oppositions, alongside examples, are listed in \Figref{fig:examples_and_diagrams}(a).

We manually labeled all pairs with their (sometimes multiple) oppositions and observe that most pairs (68\%) feature an opposition of \textit{high\slash low stature} (as in the running example), and surprisingly few pairs (7\%), one of \textit{non\hyp obscene\slash obscene}.
Due to its dominance, \Figref{fig:examples_and_diagrams}(a) further splits the \textit{high\slash low stature} opposition into 10 subtypes.

\xhdr{Main mechanism: false analogy}
Moving to a more formal analysis, we represent the running example schematically in \Figref{fig:examples_and_diagrams}(b),
while \Figref{fig:examples_and_diagrams}(c) abstracts away from the example and depicts the generic template it implements, which may be verbalized as follows.
The pair involves two entities, $x$ (God) and $x'$ (Bob Dylan), who share a crucial common property $P$ (unpredictability),
but whereas statement $P(x')$ (``Bob Dylan is unpredictable'') could potentially entail the serious headline $H(x')=h'$ (\hdln{Bob Dylan diagnosed with bipolar disorder}),
the analogous statement $P(x)$ (``God is unpredictable'') cannot entail the analogous headline $H(x)=h$ (\hdln{God  diagnosed with bipolar disorder}),
for $x$ and $x'$ are crucially opposed via one of the script oppositions of \Figref{fig:examples_and_diagrams}(a) (\textit{religion\slash no religion;} or, \textit{God,} for whom unpredictability is a sign of perfection, \vs \textit{humans,} for whom it is a sign of imperfection).
Hence, we call this mechanism \textit{false analogy.}

As the examples of \Figref{fig:examples_and_diagrams}(a) show, the analogy is never marked lexically via words such as \textit{like;} rather, it is evoked implicitly, \eg, by blending the two realms of human psychiatry and biblical lore into a single headline.
Only the satirical headline $H(x)$ itself
(red box in \Figref{fig:examples_and_diagrams}(c))
is explicit to the reader, whereas $x'$ and $P$
(and thus all the other 3 boxes)
need to be inferred.
A main advantage of our method is that it also makes $x'$ explicit and thereby facilitates inferring $P$ and thus the semantic structure that induces humor (as in \Figref{fig:examples_and_diagrams}(b)).

We emphasize that the script opposition that invalidates the logical step from $P(x)$ to $H(x)$ is not arbitrary, but must be along certain dimensions essential to human existence and contrasting ``good'' \vs ``bad'' (\Figref{fig:examples_and_diagrams}(a)).
Interestingly, in typical jokes, the ``good'' side is explicit and the ``bad'' side must be inferred,
whereas in satirical headlines, either the ``good'' or the ``bad'' side may be explicit.
And indeed, as shown by the examples of \Figref{fig:examples_and_diagrams}(a) (where the ``good'' side is marked in bold), satirical headlines differ from typical jokes in that they tend to make the ``bad'' side explicit.

\xhdr{Single \vs multiple edit operations}
A large fraction of all headlines from \textit{The Onion}---and an overwhelming fraction of those in single\hyp substitution pairs---can be analyzed with the false\hyp analogy template
(and we indeed encourage the reader to apply it to the examples of \Figref{fig:examples_and_diagrams}(a)).
Additionally, many of the pairs with two substitutions also follow this template.
\exampleNum{3} in \Figref{fig:examples_and_diagrams}(a), which plays on the opposition of the Federal Reserve being a serious institution \vs Cash4Gold being a dubious enterprise exploiting its customers, exemplifies how, whenever multiple substitutions are applied, they all need to follow the same opposition (\eg, Fed : Cash4Gold = \$85 million : \$85 = serious : dubious).

\section{Related work}
\label{sec:relwork}

The most widely accepted theory of verbal humor is the so-called \textit{General Theory of Verbal Humor} by Attardo and Raskin \shortcite{attardo1991script},
an extension of Raskin's \shortcite{raskin1985semantic} \textit{Semantic-Script Theory of Humor,}
which we summarize when discussing our findings in its context in \Secref{sec:discussion}.

Much follow-up work has built on these theories; see the excellent primer edited by Raskin \shortcite{raskin2008primer}.
Here, we focus on contributions from computer science, where most work has been on the \textbf{detection of humor} in various forms, \eg,
irony \cite{reyes2013multidimensional,wallace2015sparse},
sarcasm \cite{davidov2010semi,gonzalez2011identifying}, and
satire \cite{burfoot2009automatic,goldwasser2016understanding}, sometimes with the goal of deciding which of two texts is funnier \cite{shahaf2015inside}.
These works use documents or sentences as the smallest unit of analysis, whereas we operate at a finer granularity, analyzing the very words causing the switch from serious to funny.

Another cluster of work has considered the \textbf{generation of humor,} mostly via fixed templates such as
acronyms \cite{stock2006laughing},
puns \cite{binsted1997computational,ritchie2007practical},
two\hyp liners \cite{labutov2012humor},
or cross\hyp reference ambiguity \cite{tinholt2007computational}.

Finally, our work also relates to efforts of \textbf{constructing humor corpora} \cite{filatova2012irony,khodak2018large}.
Here, too, we increase the granularity by actively generating new data, rather than compiling humorous texts that have already been produced.
Crucially, ours is a corpus of aligned pairs, rather than individual texts, which enables entirely novel analyses that were infeasible before.

\section{Discussion and future work}
\label{sec:discussion}

\xhdr{Summary of findings}
Comparing satirical to similar\hyp but\hyp serious\hyp looking headlines within the pairs collected via \Unfun{} reveals that the humor tends to reside in the final words of satirical headlines, and particularly in noun phrases.
In order to remove the humor, players overwhelmingly replace one phrase with another; rarely do they delete phrases, and nearly never introduce new phrases.
Reversing the direction of the editing process, this implies that the most straightforward way of producing satire from a serious headline is to replace a trailing noun phrase with another noun phrase.

One may, however, not just replace any noun phrase with any other noun phrase; rather, the corresponding scripts need to be opposed along one of a few dimensions essential to the human condition and typically pitting ``good'' \vs ``bad''.
Also, the two opposing scripts need to be connected via certain subtle mechanisms, and we pointed out false analogy as one prominent mechanism.
These findings echo the predictions made by the prevailing theory of humor.
We now summarize this theory and discuss our results in its context.

\xhdr{Relation to Semantic\hyp Script Theory of Humor}
As mentioned (\Secref{sec:relwork}), the most influential theory of verbal humor has been Raskin's \shortcite{raskin1985semantic} \textit{Semantic\hyp Script Theory of Humor,} which posits
a twofold necessary condition for humorous text:
(1)~the text must be compatible with two different \textit{semantic scripts} (simply put, a semantic script is a concept together with its commonsense links to other concepts); and
(2)~the two scripts must be opposed to each other along one of a small number of dimensions.

The second criterion is key: the mere existence of two parallel compatible scripts is insufficient for humor, since this is also the case in plain, non\hyp humorous ambiguity.
Rather, one of the two scripts must be possible, the other, impossible; one, normal, the other, abnormal; or one, actual, the other, non\hyp actual.
These oppositions are abstract, and Raskin \shortcite[p.~127]{raskin1985semantic} gives several more concrete classes of opposition, which closely mirror the dimensions we empirically find in our aligned pairs (\Figref{fig:examples_and_diagrams}(a)).
Our results thus confirm the theory empirically.
But the advantages of our methodology go beyond, by letting us quantify the prevalence of each opposition.
In addition to the concrete oppositions of \Figref{fig:examples_and_diagrams}(a), we also counted how pairs distribute over the above 3 abstract oppositions, finding that most satirical headlines are of type \textit{possible\slash impossible} (64\%), followed by \textit{normal\slash abnormal} (28\%), and finally \textit{actual\slash non\hyp actual} (8\%).

In typical jokes, one of the two scripts (the so-called \textit{bona fide} interpretation) seems more likely given the text, so it is in the foreground of attention.
But in the punchline it becomes clear that the \textit{bona fide} interpretation cannot be true, causing initial confusion in the audience, followed by a search for a more appropriate interpretation, and finally surprise or relief when the actually intended, non--\textit{bona fide} script is discovered.
To enable this process on the recipient side, the theory posits that the two scripts be connected in specific ways, via the so-called
\textbf{logical mechanism,} which
resolves the tension between the two opposed scripts.

Attardo \shortcite[p.~27]{attardo2001humorous} gives a comprehensive list of 27 logical mechanisms.
While our analysis (\Secref{sec:semantic analysis}) revealed that one mechanism---\textit{false analogy}---dominates in satirical headlines, several others also occur:
\eg, in \textit{figure--ground reversal,} the real problem (the ``figure'') is left implicit, while an unimportant side effect (the ``ground'') moves into the focus of attention (\eg, \exampleNum{12} in \Figref{fig:examples_and_diagrams}(a): waterboarding, like baths, does waste water, but the real problem is ethical, not ecological).
Another common mechanism---\textit{cratylism}---plays with the assumption prevalent in puns that phonetic implies semantic similarity (\eg, \exampleNum{11} in \Figref{fig:examples_and_diagrams}(a)).

Satire is a form of art, and the examples just cited highlight that it is often the creative combination of several mechanisms that makes a headline truly funny.
Beyond the bare mechanism, the precise wording matters, too:
\eg, either \textit{16th Lassie} or \textit{17th Lassie} would suffice to make \exampleNum{6} in \Figref{fig:examples_and_diagrams}(a) funny, but the combination \textit{16th or 17th Lassie} is wittier, as it implies not only that Lassie has been played by many dogs, but also that people do not care about them, thus reinforcing the \textit{human\slash animal} opposition.

We conclude that, while satirical headlines---as opposed to typical jokes---offer little space for complex narratives, they still behave according to theories of humor.
Our contributions, however, go beyond validating these theories: the aligned corpus lets us quantify the prevalence of syntactic and semantic effects at play and reveals that the dominant logical mechanism in satirical headlines is false analogy.

\xhdr{Satirical\hyp headline generation}
This points to a way of generating satirical headlines by implementing the false\hyp analogy template of \Figref{fig:examples_and_diagrams}(c):
pick an entity $x$
(\eg, Pepsi)
and a central property $P(x)$ of $x$
(\eg, ``Pepsi is a popular drink'');
then pick another entity $x'$ for which $P(x')$ also holds, but which is opposed to $x$ along one of the axes of \Figref{fig:examples_and_diagrams}(a)
(\eg, Bordeaux wine, which is in a \textit{high\slash low stature} [\textit{sublime\slash mundane}] opposition to Pepsi);
and finally generate a headline $H(x')$ based on $P(x')$
(\eg, \hdln{2018 Bordeaux vintage benefits from outstanding grape harvest})
which cannot be seriously formulated for $x$ instead $x'$, due to the opposition, yielding the satirical $H(x)$
(\eg, \hdln{2018 Pepsi vintage benefits from outstanding high\hyp fructose corn harvest,} where we analogously replaced \hdln{grape} with \hdln{high\hyp fructose corn,} \cf \Secref{sec:semantic analysis}).
The subtitle of the present paper was also generated this way.

Most humans are unaware of the logical templates underlying satire,
while machines have difficulties finding entity pairs opposed in specific ways and formulating pithy headline text.
We hence see promise in a hybrid system for coupling the respective strengths of humans and machines, where the machine guides the human through the template instantiation process while relying on the human for operations such as finding appropriate entities for substitution \etc

\xhdr{Human perception of satirical \vs serious news}
Recall that in task~2 (\Secref{sec:game}), players also rate unmodified satirical and serious headlines $g$ with respect to how likely they consider them to be serious.
\Tabref{tbl:confusion_matrix} shows that, although players are generally good at distinguishing satire from real news, they do make mistakes:
10\% of serious headlines are consistently misclassified as satirical (\eg, \hdln{Schlitz returns, drums up nostalgic drinkers}),
and 8\% of satirical headlines, as serious (\eg, \hdln{Baltimore looking for safer city to host Super Bowl parade}).
Studying these misunderstood headlines can yield interesting insights into how readers process news, especially in an age where ``fake news'' is becoming a ubiquitous scourge.
We leave this analysis for future work.

\xhdr{Beyond humor}
The mechanism underlying \Unfun{} defines a general procedure for identifying the essential portion of a text that causes the text to have a certain property.
In our case, this property is humor, but when asking players instead to remove the rudeness, sexism, euphemism, hyperbole, \etc, from a given piece of text,
we obtain a scalable way of collecting fine\hyp grained supervised examples for better understanding these ways of speaking linguistically.

\section{Conclusion}
Humor is key to human cognition and holds questions and promise for advancing artificial intelligence.
We focus on the humorous genre of satirical news headlines
and present \Unfun, an online game for collecting pairs of satirical and similar\hyp but\hyp serious\hyp looking headlines,
which precisely reveal the humor\hyp carrying words and the semantic structure in satirical news headlines.
We hope that future work will build on these initial results,
as well as on the dataset that we publish with this paper \cite{github},
in order to make further progress on understanding satire
and, more generally, the role of humor in intelligence.

\end{document}